\documentclass[10pt, conference, compsocconf]{IEEEtran}
\IEEEoverridecommandlockouts
\usepackage{cite}
\usepackage{amsmath,amssymb,amsfonts}
\usepackage{algorithmic}
\usepackage{graphicx}
\usepackage{subfigure}
\usepackage{epstopdf}
\usepackage{textcomp}
\usepackage{xcolor}
\def\BibTeX{{\rm B\kern-.05em{\sc i\kern-.025em b}\kern-.08em
    T\kern-.1667em\lower.7ex\hbox{E}\kern-.125emX}}    
\newcommand{\fig}[1]{Figure \ref{#1}}
\newcommand{\eq}[1]{Eq.\ref{#1}}
\newcommand{\tab}[1]{Table \ref{#1}}
\newcommand{\tabincell}[2]{\begin{tabular}{@{}#1@{}}#2\end{tabular}}

\renewcommand{\IEEEbibitemsep}{0pt plus 0.5pt}
\makeatletter
\IEEEtriggercmd{\reset@font\normalfont\fontsize{8.4pt}{8.5pt}\selectfont}
\makeatother
\IEEEtriggeratref{1}

\begin{document}
\title{Patch Aggregator for Scene Text Script Identification\\
\thanks{\textsuperscript{\#}Both the authors contributed equally.}
\thanks{\textsuperscript{*}Corresponding author.}
}

\author{\IEEEauthorblockN{Changxu Cheng\textsuperscript{\#}, Qiuhui Huang\textsuperscript{\#}, Xiang Bai\textsuperscript{*}, Bin Feng, Wenyu Liu}
\IEEEauthorblockA{\textit{School of Electronic Information and Communication} \\
\textit{Huazhong University of Science and Technology, Wuhan, China}\\
\{cxcheng, xbai, liuwy\}@hust.edu.cn, lairehuang@gmail.com, frobby@163.com}
}

\maketitle

\begin{abstract}
Script identification in the wild is of great importance in a multi-lingual robust-reading system. The scripts deriving from the same language family share a large set of characters, which makes script identification a fine-grained classification problem. Most existing methods make efforts to learn a single representation that combines the local features by making a weighted average or other clustering methods, which may reduce the discriminatory power of some important parts in each script for the interference of redundant features. In this paper, we present a novel module named Patch Aggregator (PA), which learns a more discriminative representation for script identification by taking into account the prediction scores of local patches. Specifically, we design a CNN-based method consisting of a standard CNN classifier and a PA module. Experiments demonstrate that the proposed PA module brings significant performance improvements over the baseline CNN model, achieving the state-of-the-art results on three benchmark datasets for script identification: SIW-13, CVSI 2015 and RRC-MLT 2017.

\end{abstract}


\IEEEpeerreviewmaketitle

\section{Introduction}\label{intro}
Script identification is to predict the script of a given text image, having played a more and more important role in multilingual systems nowadays. Under many circumstances, it acts as a prerequisite to decide which language model to use for further text detection or recognition.

Earlier works conducted on document~\cite{docSI1,docSI2,docSI3,docSI4}, handwritten text~\cite{handSI1,handSI2} and video overlaid-text~\cite{ 1st_VTSI ,videoSI1,videoSI2} where texts hold regular layout and simple background have achieved great performance. But when it comes to scene text script identification which extends the application to more fields like scene understanding~\cite{sceneUnderstand}, additional challenges emerge, like the complex background, various text styles and diverse noise, etc. Our work focuses on scene text, taking on challenges as follows:
\begin{itemize}
\item Some scripts have relatively subtle differences, e.g., Russian and English, which share a large set of characters. Distinguishing them is exactly a fine-grained classification problem requiring discriminative features. 
\item Cropped text images have arbitrary aspect ratios, making it necessary to find an effective way to feed them into the model in the batch-based training phase.
\end{itemize}

\begin{figure}[htbp]\label{fig}
\setlength{\abovecaptionskip}{0.cm}
\setlength{\belowcaptionskip}{-0.cm}
\centering

\subfigure[]{
\centering
\includegraphics[width=8.5cm]{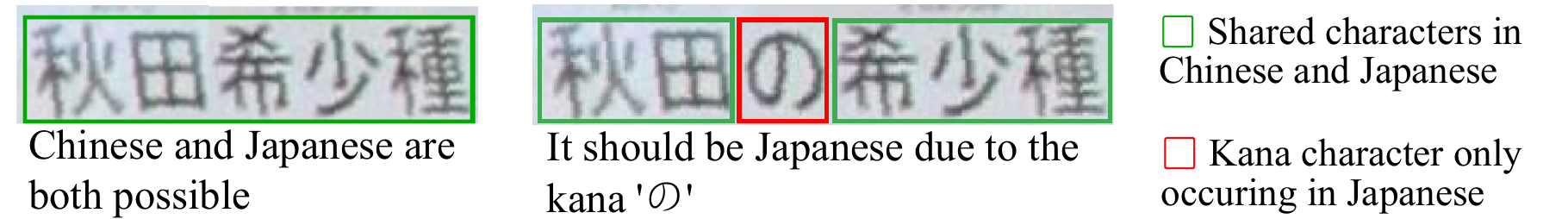}
\label{fig1a}}

\subfigure[]{
\centering
\includegraphics[width=8.5cm]{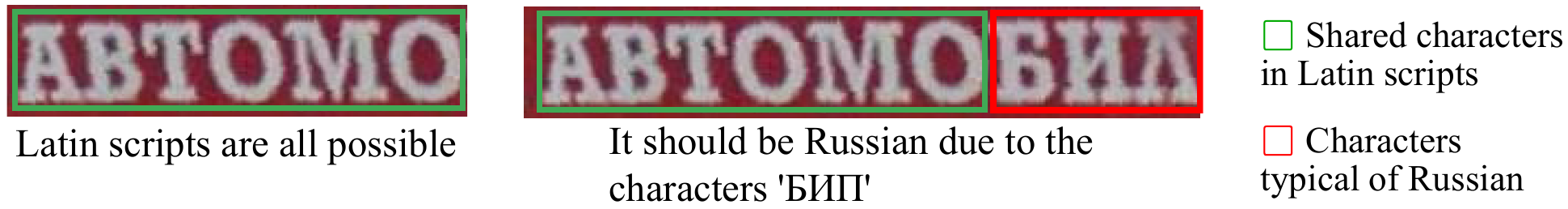}
\label{fig1b}}

\centering
\caption{Examples of SIW-13 illustrate the importance of local discriminative parts. Characters in the red bounding box are discriminative, while those in the green one occur in several scripts. (a) The left text line only consisting of shared characters may be Chinese or Japanese, which is ambiguous. As for the right with just a kana character added, we are sure it is Japanese. (b) The left can be any type of Latin, but the right is definitely Russian due to the discriminative characters bounded by the red box.}
\end{figure}

The first challenge is crucial in script identification where the bottleneck mainly comes from the scripts of the same family sharing some common characters. Hence, local discriminative features are always paid much attention to. Almost all the works focus on collecting critical features without suppressing the redundant features acting as noise. 
Some works~\cite{discluster,FGSI,bag} adopted clustering on deep convolutional features for critical descriptors. There were multi-stage training process and great computation due to clustering. 
Inspired by Siamese network~\cite{bromley1994signature}, Gomez et al.~\cite{improveFGSI} proposed an improved patch-based method containing an ensemble of identical nets to learn discriminative stroke-part representations. 
Mei et al.~\cite{CRNN} adopted Convolutional Recurrent Neural Networks~\cite{shi2017end} to extract the image representation and spatial dependency which is discriminative in spite of sharing characters.
Fujii et al.~\cite{seq2label} use Encoder and Summarizer to get local features and fuse them to a single summary by attention mechanism~\cite{attention} to reflect the importance of different patches.
Ankan et al.~\cite{attLSTM} proposed an attention-based Convolutional-LSTM network, analyzing features globally and locally which is popular in fine-grained classification~\cite{dfl,macnn}.

\begin{figure}[htbp]
\begin{center}
\includegraphics[scale=0.75]{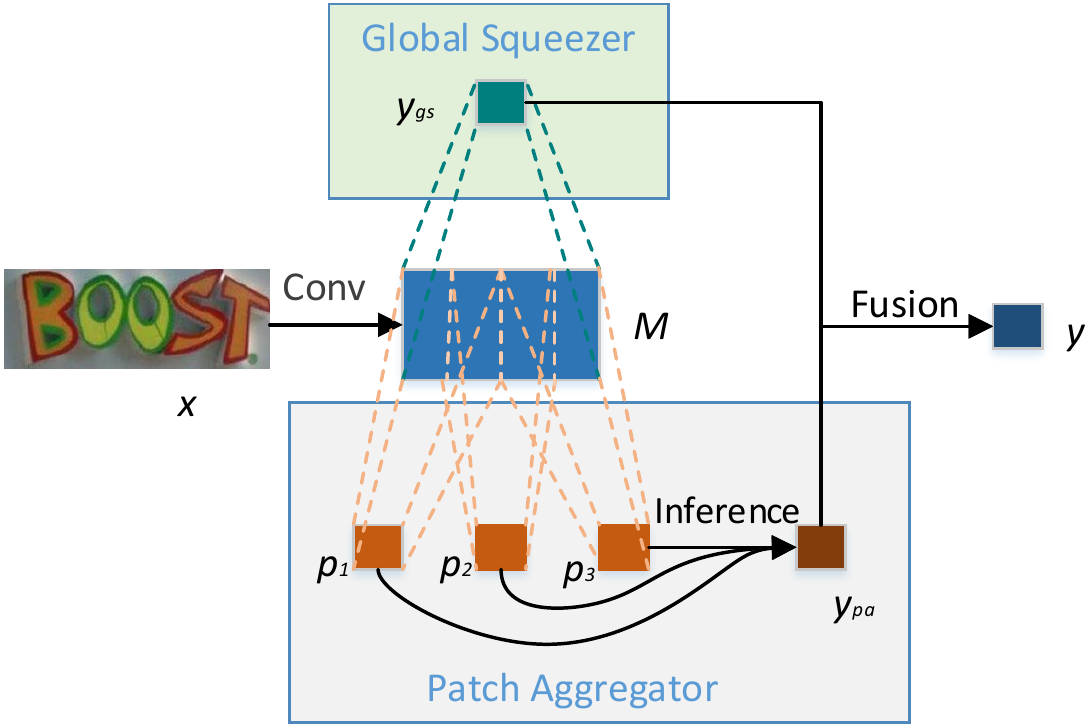}
\end{center}
\caption{Overview of our proposed method. Basic features $M$ is first extracted  from input $x$. Then Global Squeezer makes a general prediction $y_{\mathrm{gs}}$, and meanwhile Patch Aggregator mines patch-level prediction scores $p_1, p_2, ..., p_n$ to construct a representation for the later prediction $y_{\mathrm{pa}}$ with dropping redundant features. Finally $y_{\mathrm{gs}}$ and $y_{\mathrm{pa}}$ are dynamically fused to output $y$. It can be trained end-to-end in one stage.}
\label{fig2}
\end{figure}

Although having made great progress, the works above taking features of all patches into account suffer from a fatal issue that the domination of the discriminative features could be reduced by other weak-discrimination features. Especially, a text line attached to a specific script can consist of many characters belonging to the intersection of several scripts, making a model prone to suffer from redundant features.
As shown in \fig{fig1a}, the text line in the left only consisting of shared characters can be either Chinese or Japanese. However, the right one with only one character added is definitely Japanese, which shows the great power of the discriminative features. But the existing works cannot make good use of the features. For example, if we average all the character patches with weights, the power of the discriminative patches will be reduced by a much larger number of shared character patches.
A similar case is in \fig{fig1b} where the few Russian-specific characters on the right are critical. 

The discriminative part is expected to be dominant even if with smaller quantity. Here we propose Patch Aggregator (PA) to learn and aggregate the local features.
PA makes patch-level predictions as an explicit representation, from which we can know what scripts the patches of a given image could be. After that, by simply max-pooling the predicted probability distributions, the relation between the input image and every script is obvious. This is a low-dimensional but important discriminative feature representation. Based on this, a simple linear classifier will make a local-level prediction about the whole image.
For example, the right image in \fig{fig1a} contains patches attached to two scripts, i.e., Chinese and Japanese, which form the low-dimensional discriminative features. PA will predict that it is more likely to be Japanese if both Chinese and Japanese-specific characters occur in an image. But when no Japanese-specific character occurs, as in the left in \fig{fig1a}, PA will infer it as Chinese. The process can be learned well in the training stage.

As for the problem of arbitrary aspect ratios, 
recent methods with good performance take densely cropped image patches with fixed size as input~\cite{FGSI,improveFGSI,bag,attLSTM}.
They also employ data augmentation somehow, but they suffered from the following three issues.
Firstly, A cropped image patch may bring noise caused by sudden breaking off. And the feature extractor cannot catch its surroundings in other patches, which limits the feature representation for losing the holistic context messages.
Secondly, heavy redundancy of overlapped patches could lead to much repeated computation, pulling down the efficiency during test phase.
Thirdly, the samples with larger aspect ratios in some scripts could make more cropped patches which may cause great data imbalance, disturbing the training to some degree.
Hence, our input prefers full-size images to cropped patches. Shi et al.~\cite{wildSI} designs a spatially-sensitive pooling layer by pooling horizontally on the intermediate feature map so that the width of the input image can be flexible. We adopt pooling strategy to solve the problem too, but our pooling process intends to keep more useful information and be more interpretable.

In this work, we employ an end-to-end CNN-based method consisting of a standard CNN classifier called Global Squeezer (GS) and a PA module, as shown in \fig{fig2}.
In training phase, we design a novel loss called softermax loss to take patch-level predictions under the weak supervision of the ground truth label in PA, since the label of the whole text image sometimes cannot imply the exact classes of patches for the characters sharing of some scripts. All other predictions are supervised by softmax loss. 
Succinctly, the main contributions of this paper are as follows:
\begin{enumerate}
\item We propose PA to aggregate patch-level predictions to learn a discriminative representation, which has high interpretability. PA along with GS can process images with arbitrary aspect ratios in a simple but effective way.
\item We design softermax loss to accomplish patch-level weak supervision on local predictions with image-level label.
\item Experiments are conducted on three public datasets, i.e., SIW-13~\cite{improvedSI}, CVSI2015~\cite{CVSI} and RRC-MLT2017~\cite{MLT}, and achieve state-of-the-art performance.
\end{enumerate}

\begin{figure*}[htbp]
\begin{center}
\includegraphics[scale=0.65]{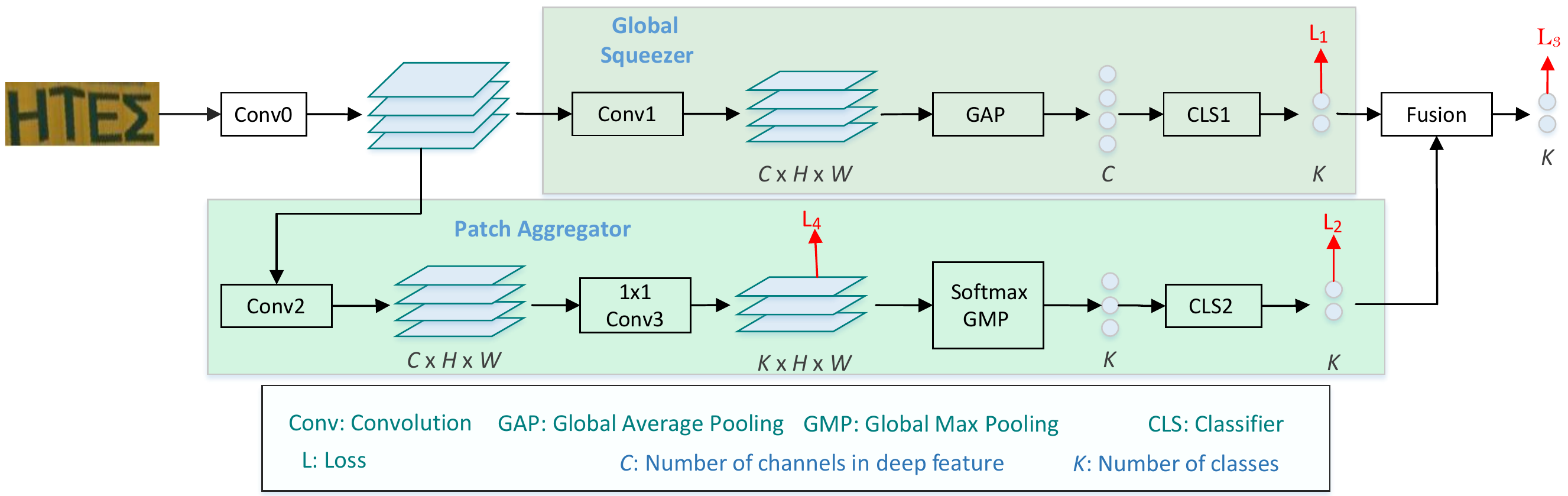}
\end{center}
\caption{Details of our method. The full-size image is fed into the shared convolution module Conv0 at first. Conv1 and Conv2 are specially designed for the two modules. The upper module GS uses Global Average Pooling (GAP) to squeeze the features per channel for general prediction. Meanwhile, the lower module PA employs $1 \times 1$ convolution and softmax to make patch-level prediction for intermediate supervision. Global Max Pooling (GMP) applied to the patch-level prediction scores of every class got before implies what potential scripts the input could belong to. Subsequently the fully connected layer (CLS2) makes fine-grained prediction. GMP and CLS2 show the inference process to mine the discriminative features and consider the context. Finally we take adaptive weighted sum of the two modules. It is supervised by 4 losses to ensure output in expectation in training phase. 
\label{fig3}}
\vspace{-0.4cm}
\end{figure*}

\section{Methodology}\label{methodology}

Our proposed method can perform script identification simply and effectively. Convolutional operation is directly imposed on a full-size image instead of cropped patches. 

An overview of our method is shown in Figure \ref{fig2} and the detail is in \fig{fig3}. The patch mentioned here is a single pixel of specific deep feature map with a proper size of receptive field.
A shared convolutional structure acts as a basic feature extractor in the framework, followed by two modules called Global Squeezer (GS) and Patch Aggregator (PA) respectively. GS aims to squeeze the holistic representation while PA makes predictions over local features and aggregates them by inference which is able to make full use of the discriminative features. Finally, we fuse them dynamically in a learnable way. The entire network can be trained end-to-end in one stage.

\subsection{Global Squeezer}%\label{AA}
Once we get the basic feature by the shared convolutional structure, Global Squeezer (GS), a common classifier, makes a global prediction. Firstly a tiny convolutional structure gets the feature map $\mathbb{M}$, which meets the demands of globally squeezing, including receptive field and dimension. Subsequently we squeeze $\mathbb{M}$ (of size $\mathrm{C}\times{\mathrm{H}}\times{\mathrm{W}}$) across spatial dimensions by Global Average Pooling (GAP) to get a channel-wise global descriptor $\mathrm{Z}\in \displaystyle{R^\mathrm{C}}$, where $\mathrm{C}$ is the number of channels. This can be described as \eq{avgpol}.
{\setlength\abovedisplayskip{1pt}
\setlength\belowdisplayskip{1pt}
\begin{equation}
\label{avgpol}
\displaystyle{\mathrm{Z}_i = \frac{1}{\mathrm{W}{\mathrm{H}}}{\sum_{w=1}^{\mathrm{W}}\sum_{h=1}^{\mathrm{H}}\mathbb{M}( i , w , h )}\ \ i=1,2,3,\cdots,\mathrm{C}}
\end{equation}}

GAP squeezes holistic feature representation across each channel, since a convolutional feature channel often corresponds to a certain type of visual pattern~\cite{zhang2016picking}. Then the holistic representation is fed into a linear classifier to get the global prediction scores $y_{\mathrm{gs}}$ over $\mathrm{K}$ classes.

\subsection{Patch Aggregator}\label{BB}

It’s far from being sufficient to learn discriminative features by just making prediction from a single global perspective as GS does. Specifically, attention mechanism~\cite{attention} has been widely applied to accomplish discriminative learning extensively. However, it seems not so valid as for script identification due to the effect caused by redundant features. Here we specify the novel Patch Aggregator (PA) which learns and employs the discriminative features better.

PA starts with the same tiny convolutional structure as in GS to make pixels in deep feature map of proper receptive fields which result in precise patch-level scores implemented by $1\times 1$ convolution. Then softmax function converts the scores to probability distributions $\mathbb{Y}_\mathrm{inter}$ (of size $\mathrm{K} \times \mathrm{H} \times \mathrm{W}$ across the $\mathrm{K}$ classes). This step goes under a special intermediate supervision discussed in \ref{softer} during training. The patch-level scores actually act as high-level semantic features where the discriminative representation can be extracted .

Taking account of the impairment caused by redundant features, we adopt Global Max Pooling (GMP) when we aggregate the prediction scores of patches to pick the most remarkable response per class, which is highly interpretable. The process can be described as \eq{maxpol}.
\begin{equation}
\label{maxpol}
\displaystyle{\mathrm{q}_i = \max\limits_{w,h}\left\{\mathrm{p}(i,w,h)\right\} \ i=1,2,\cdots,\mathrm{K}}
\end{equation}
where $\mathrm{p}\left(i,w,h\right)$ is the score of the patch in position $(w,h)$ corresponding to the $i_{th}$ class. After we pick out the maximum of $\mathrm{p}$ per class, $\mathrm{q}$ reflects the likelihood of the given image appertaining to every class, thus we can know which scripts the components of the input image could belong to. Then a two-layer linear classifier gets the scores $y_{\mathrm{pa}}$ over $\mathrm{K}$ classes from local perspective.

Visualization of the behaviour in the module is available in \ref{visualization}.

\subsection{Fusion}

To combine the outputs of the above two modules adaptively, we adopt dynamic
weighted fusion. The weight of global output $y_{\mathrm{gs}}$ just depends on $y_{\mathrm{gs}}$ itself somehow, denoted as $\gamma$. 
Then the weight of $y_{\mathrm{pa}}$ is the complement $\left(1-\gamma \right)$. The fusion process can be shown in \eq{fusion1} and \eq{fusion2}. \eq{fusion1} show the mapping process and $\sigma$ is the sigmoid function. $W$ and $b$ are trainable parameters of linear layers.

\begin{equation}
\setlength\abovedisplayskip{0pt}
\setlength\belowdisplayskip{0pt}
\vspace{-0.5cm}
\label{fusion1}
\displaystyle{\gamma = \sigma(Wy_{\mathrm{gs}} + b)}
\vspace{0.2cm}
\end{equation}

\begin{equation}
\setlength\abovedisplayskip{0pt}
\setlength\belowdisplayskip{0pt}
\label{fusion2}
\displaystyle{y = \gamma y_{\mathrm{gs}} + (1-\gamma)y_{\mathrm{pa}}}
\vspace{0.2cm}
\end{equation}

\subsection{Loss Functions}

In the training stage, the proposed network is optimized by four losses--- $\mathbb{L}_1$ $\mathbb{L}_2$ $\mathbb{L}_3$ and $\mathbb{L}_4$ as shown in Figure \ref{fig3}, to make sure the network can work within our expectation.

GS and PA are both under supervision with $\mathbb{L}_1$ and $\mathbb{L}_2$ respectively to make sure they really learn well. $\mathbb{L}_3$ is devised for the final decisive output $y$ which determines the performance of the model, holding a relatively higher weight. The three losses all use softmax loss based on the ground truth labels.

The loss $\mathbb{L}_4$ is designed for the intermediate supervision as have been mentioned in section \ref{BB}. Since the categories of some patches cannot simply rely on the image-level label for the characters-sharing issue, the challenge turns out that the image-level label is not sufficient to supervise patch-level scores if we directly use softmax loss. Thus we propose the novel softermax loss to deal with the problem.

\subsubsection{Softermax Loss}\label{softer}
Classical softmax loss pushes the model to output a much greater probability on the ground truth(GT) class than others. It makes the model excessively confident in GT, which is inappropriate for patch-level prediction for scripts confusion of some characters. To relieve the extreme and fully learn discriminative features in patch-level, we make the loss softer for a single patch, which can be formulated as in \eq{softermax loss}.
{\setlength\abovedisplayskip{1pt}
\begin{equation}
\label{softermax loss}
\mathbb{L}_\mathrm{softer} = -\log\left(\frac{\sum\limits_{j=1}^{k}e^{x_{m_j}}}{\sum\limits_{i=1}^{\mathrm{K}}e^{x_i}}\right) 
\end{equation}}
where $x_i$ is the score about the i-\textit{th} category at a specific location got by $1\times1$ convolution, and $x_{m_1}, \cdots ,x_{m_k}$ are the top-$k$ elements of $\left\{x_i \right\}_{i=1}^{\mathrm{K}}$($k$ is a hyperparameter). $\mathbb{L}_\mathrm{softer}$ prompt the top $k$ probabilities to be as great as possible, alleviating the extreme of softmax loss up to a point. But it is unsupervised learning to just adopt the softermax loss, leaving the model prone to fall into local optimum. 

Hence we couple the softmax and softermax loss to get a trade-off. The loss for an image is averaged over  its $\mathrm{N}$ patches, which is $\mathbb{L}_4$ shown in \eq{L4}, where $\lambda$ determines how softer \eq{L4} could be, and $\mathbb{L}_\mathrm{CE}$ is the softmax loss supervised simply by the label of the input image.
{\setlength\abovedisplayskip{1pt}
\setlength\belowdisplayskip{1pt}
\begin{equation}
\label{L4}
\mathbb{L}_4 = \frac{1}{\mathrm{N}}\sum\limits_{n=1}^{\mathrm{N}} \left(\lambda\mathbb{L}_\mathrm{softer}^{(n)} + \left(1-\lambda\right)\mathbb{L}_\mathrm{CE}^{(n)} \right)
\end{equation}}
{\setlength\abovedisplayskip{1pt}
\setlength\belowdisplayskip{1pt}
\begin{equation}
\label{L}
\mathbb{L} = \eta_1\mathbb{L}_1+\eta_2\mathbb{L}_2 +\eta_3\mathbb{L}_3 + \eta_4\mathbb{L}_4
\end{equation}}

\begin{table}[htb]
\setlength{\abovecaptionskip}{0.cm}
\setlength{\belowcaptionskip}{-0.0cm}
\caption{Grouping Resizing with height=32 for SIW-13.}
\begin{center}
\begin{tabular}{c|c|c|c|c}
\hline 
Original Aspect Ratio Range & (0,3) & [3,6) & [6,12] & (12,) \\ 
\hline 
New Aspect Ratio & 2 & 4 & 8 & 16 \\ 
\hline 
\end{tabular}
\label{SIWgroup}
\end{center}
\vspace{-0.5cm}
\end{table}

During training, the above losses contribute to the total loss by weights $\{\eta_i\}_{i=1}^4$, as shown in \eq{L}.

\section{Experiment}
We conduct experiments on three public datasets for script identification.\, \textbf{SIW-13}~\cite{improvedSI} is officially split into 9,791 training and 6,500 test images of 13 scripts.
\, \textbf{CVSI2015}~\cite{CVSI} is released for the ICDAR 2015 Competition on Video Script Identification, containing text line images of 10 Indian scripts.
\, \textbf{RRC-MLT2017}~\cite{MLT} is released for ICDAR 2017 Competition on MLT-Task2, comprising 68,613 training, 16,255 validation and 97,619 test cropped images.
This dataset holds an extremely imbalanced distribution among 7 scripts and especially tilts to Latin. There exists some multi-oriented and curved texts which make it more challenging.

\subsection{Implementation Details}
In terms of the diverse aspect ratios of the dataset images, we group every image by its aspect ratio and resize it to a fixed size determined by the group it belongs. The short side of all images are set to 32. Then we can train them with batches efficiently. The number of groups is determined by the dataset. To be clear, Table \ref{SIWgroup} shows the grouping resizing in SIW-13. For example, if an image has an aspect ratio of 3.5, we should resize it to size 32x128 where 32 is the fixed height. The same trick is used on CVSI2015 and RRC-MLT2017.

We also exploit some data augmentation like changing contrast, adding random noise, slightly cropping and making perspective transform to make full use of training data. Image data is normalized in range $[-1,1]$ uniformly.

Our basic architecture uses VGG~\cite{VGG}-style stacking, i.e., 3x3 convolution with 1 padding followed by Batch Normalization~\cite{ioffe2015batch} and ReLU. 
More details are shown in Table \ref{archi1} for SIW-13 and CVSI2015, where Module 1-6 are the shared convolutional part and GS stands at the left while PA is at the right. Note that "1-6" means the first 6 modules have the same structure shown in the right but with different number of filters and parameters. We use kaiming normalization~\cite{kaiming_normal} to initialize it.
As for RRC-MLT2017, we take the convolutional part of VGG16~\cite{VGG} pre-trained in ImageNet as the backbone due to the much more complex images.
The design guarantees enough receptive field for patch-level prediction sores.

\begin{table*}[htb]
\setlength{\abovecaptionskip}{0.cm}
\setlength{\belowcaptionskip}{-0.1cm}
\caption{Architecture of our framework for SIW-13 and CVSI2015.}\label{archi1}
\begin{center}
\begin{tabular}{|c|c|c|}
\hline 
No. of Module & \multicolumn{2}{c|}{Configuration} \\ 
\hline 
1-6 & \multicolumn{2}{c|}{\tabincell{c}{\textbf{Conv} kernel:$(64,128,256)\times{3}\times{3}$, stride:$1\times{1}$, padding:$1\times{1}$ \\ \textbf{BatchNorm} \qquad \textbf{ReLU}\qquad(\textit{3 kind of convolutional kernels for Module 1-2,3-4,5-6 respectively})}} \\ 
\hline 
\textit{Pooling} & \multicolumn{2}{c|}{\tabincell{c}{\textbf{MaxPooling} kernel:$\times{2}\times{2}$, stride:$2\times{2}$ \qquad (\textit{placed after Module 2,4,6})}}\\
\hline
7 & \tabincell{c}{\tabincell{c}{\textbf{Conv} kernel:$512\times{3}\times{3}$, stride:$1\times{1}$, padding:$0\times{1}$\\\textbf{BatchNorm}\qquad \textbf{ReLU}}} & \tabincell{c}{\tabincell{c}{\textbf{Conv} kernel:$512\times{3}\times{3}$, stride:$1\times{1}$, padding:$0\times{1}$\\\textbf{BatchNorm}\qquad \textbf{ReLU}}} \\ 
\hline 
8 & \tabincell{c}{\textbf{Conv} kernel:$512\times{2}\times{3}$, stride:$1\times{2}$, padding:$0\times{1}$\\
\textbf{BatchNorm}\qquad \textbf{ReLU}} & \tabincell{c}{\textbf{Conv} kernel:$512\times{2}\times{3}$, stride:$1\times{2}$, padding:$0\times{1}$\\\textbf{BatchNorm}\qquad \textbf{ReLU}} \\ 
\hline 
9 & \tabincell{c}{\textbf{Linear}:512 \qquad \textbf{ReLU} \qquad \textbf{Dropout(0.3)}} & \tabincell{c}{\textbf{Conv} kernel:$128\times{1}\times{1}$, stride:1, padding:0\\ \textbf{BatchNorm}\qquad \textbf{ReLU}} \\ 
\hline 
10 & \textbf{Linear}:$K$ & \textbf{Conv} kernel:$K\times{1}\times{1}$, stride:1, padding:0 \\ 
\hline 
11 & - & \tabincell{c}{\textbf{Linear}:32 \qquad \textbf{ReLU}} \\ 
\hline 
12 & - & \textbf{Linear}:$K$ \\ 
\hline 
Fusion & \multicolumn{2}{c|}{\tabincell{c}{\textbf{Linear}:1 \qquad \textbf{Sigmoid}}} \\ 
\hline 
\end{tabular} 
\end{center}
\vspace*{-0.6cm}
\end{table*}

\begin{table}[htb]
\setlength{\abovecaptionskip}{0.cm}
\setlength{\belowcaptionskip}{-0.1cm}
\caption{Results of our method on SIW-13, CVSI2015, RRC-MLT2017, as well as some other methods to be compared with.}
\begin{center}
\begin{tabular}{l|ccc}
\hline 
Method & SIW-13 & CVSI2015 & RRC-MLT2017 \\ 
\hline 
Shi~\cite{wildSI} & 88.0 & 96.69 & - \\ 
Shi~\cite{improvedSI} & 89.4 & 94.30 & - \\ 
Gomez~\cite{improveFGSI} & 94.8 & 97.20 & - \\ 
Nicolaou~\cite{nicolaou2016visual} & 83.7 & 98.18 & - \\ 
Mei~\cite{CRNN} & 92.75 & 94.20 & - \\ 
Bhunia~\cite{attLSTM} & 96.5 & 97.75 & - \\ 
Zdenek~\cite{bag} & 92.88 & 97.11 & - \\ 
Patel~\cite{patel2018e2e} & - & - & 88.54 \\
\hline 
ours & \textbf{97.3} & \textbf{98.60} & \textbf{89.42} \\ 
\hline 
\end{tabular}
\label{results}
\end{center}
\vspace{-0.4cm}
\end{table}

\begin{table}[htbp]
\setlength{\abovecaptionskip}{0.cm}
\setlength{\belowcaptionskip}{-0.1cm}
\caption{Accuracies for all script types, Model size (MB) and test speed (ms per img) on SIW-13}\label{scriptsRes}
\begin{tabular}{l|cccccc}
\hline
Script & Zdenek~\cite{bag} & Mei~\cite{CRNN} & Gomez~\cite{improveFGSI} & Bhunia~\cite{attLSTM} & ours  \\
\hline 
Avg & 92.88 & 92.75 & 94.8  & 96.5 & \textbf{97.3} \\
\hline 
Ara & 97.0 & 96.2  & 98.0  & \textbf{99.0} & 98.6  \\
Cam & 96.8 & 93.4  & \textbf{99.2} & 99.0 & 98.6  \\
Chi & 91.3 & 94.0  & 88.4  & 92.0 & \textbf{95.6}  \\
Eng & 80.5 & 83.6  & 97.0  & \textbf{98.0} & 94.0    \\
Gre & 84.6 & 89.4  & 99.8  & \textbf{100.0} & 96.4  \\
Heb & 94.6 & 93.8  & 96.2  & \textbf{99.0} & 96.8  \\
Jap & 93.4 & 91.8  & 92.6  & \textbf{98.0} & 95.2  \\
Kan & 94.7 & 91.8  & 88.6  & 92.0 & \textbf{98.0}  \\
Kor & 97.5 & 95.6  & 89.4  & 93.0 & \textbf{99.6} \\
Mon & 97.7 & 97.0  & 94.6  & 98.0 & \textbf{98.8} \\
Rus & 82.1 & 87.0  & \textbf{95.0} & 93.0 & 94.8 \\
Tha & 97.2 & 93.6  & 94.8  & 95.0 & \textbf{98.4} \\
Tib & 99.2 & 98.6  & 98.2  & 97.0 & \textbf{99.8}  \\
\hline
Size & - & -  & 24  & 12 & 26.7  \\
Speed & 60 & 92  & 13  & 85 & 2.5  \\
\hline
\end{tabular}
\vspace{-0.5cm}
\end{table}

In the experiments we have used PyTorch~\cite{pytorch} for deep learning acceleration. During training, hyper parameters for \eq{softermax loss}, \eq{L4} and \eq{L} are: $k=3$, $\lambda = 0.4$, [$\eta _{1,2,3,4}$] = [0.1, 0.1, 1.0, 0.1], which can lead to the best accuracy. The batch size is 16. Stochastic gradient decent (SGD) is used for optimization with momentum and weight decay set to 0.9 and 1e-4 respectively. Learning rate starts with 0.1 and will decay by 0.3 if the training loss stop falling for a while. Every time it is lower than 8e-5, we reset it to 0.01 and going on training until the default epoch (500 for SIW-13 and CVSI2015, 100 for RRC-MLT2017) is reached.
We conduct our experiment on an Nvidia Geforce GTX GPU with 10.9GB memory, one Intel(R) Xeon(R) CPU E5-2637 v4 @ 3.50GHz and 64GB RAM. The training time is around 5 hours.

\begin{table}[htbp]
\setlength{\abovecaptionskip}{0.cm}
\setlength{\belowcaptionskip}{-0.1cm}
\caption{Contribution of the proposed branches on SIW-13. }\label{branch}
\begin{center}
\begin{tabular}{l|ccccc}
\hline 
Script & GS & GS+GS & GS+GMP & PA & GS+PA \\
\hline 
Avg & 96.2 & 96.3 & 96.5 & 94.5 & \textbf{97.3} \\
\hline
Ara & 98.2 & 98.4 & 97.8 & 96.8 & \textbf{98.6} \\

Cam & 95.6 & 96.8 & 97.2 & 94.8 & \textbf{98.6} \\
 
Chi & 94.8 & 95.0 & 94.6 & 94.4 & \textbf{95.6} \\

Eng & 91.6 & 90.4 & 92.4 & 89.8 & \textbf{94.0} \\

Gre & 95.0 & \textbf{96.6} & 96.4 & 92.2 & 96.4 \\

Heb & 96.4 & 96.4 & \textbf{97.2} & 95.2 & 96.8 \\

Jap & \textbf{95.6} & 95.6 & 95.6 & 92.4 & 95.2 \\

Kan & 96.2 & 97.4 & 98.0 & 96.4 & \textbf{98.0} \\

Kor & 98.6 & 98.0 & 98.4 & 96.8 & \textbf{99.6} \\
 
Mon & 98.6 & 98.0 & 98.4 & 98.2 & \textbf{98.8} \\

Rus & 92.8 & 92.2 & 90.8 & 87.0 & \textbf{94.8} \\

Tha  & 98.0 & 98.0 & 98.2 & 95.8 & \textbf{98.4} \\

Tib & 99.6 & 99.4 & 99.8 & 99.2 & \textbf{99.8} \\
\hline
\end{tabular}
\end{center}
\vspace{-0.5cm}
\end{table}

\begin{table}[htb]
\setlength{\abovecaptionskip}{0.cm}
\setlength{\belowcaptionskip}{-0.1cm}
\caption{Effects of the intermediate supervision and Softmax loss in PA on SIW-13.}\label{supervision}
\begin{center}
\begin{tabular}{l|c|c|c}
\hline 
$\mathbb{L}$ & without $\mathbb{L}_4$ & with $\mathbb{L}_4(\lambda =0)$ & with $\mathbb{L}_4(\lambda =0.4)$ \\ 
\hline 
$Acc$ & 96.3 & 96.8 & \textbf{97.3} \\ 
\hline 
\end{tabular} 
\end{center}
\vspace{-0.6cm}
\end{table}

\subsection{Results}
The results on SIW-13, CVSI2015 and RRC-MLT2017 are displayed in \tab{results} and \tab{scriptsRes}.

\begin{figure*}[htbp]
\centerline{
\subfigure[]{\includegraphics[scale=0.55]{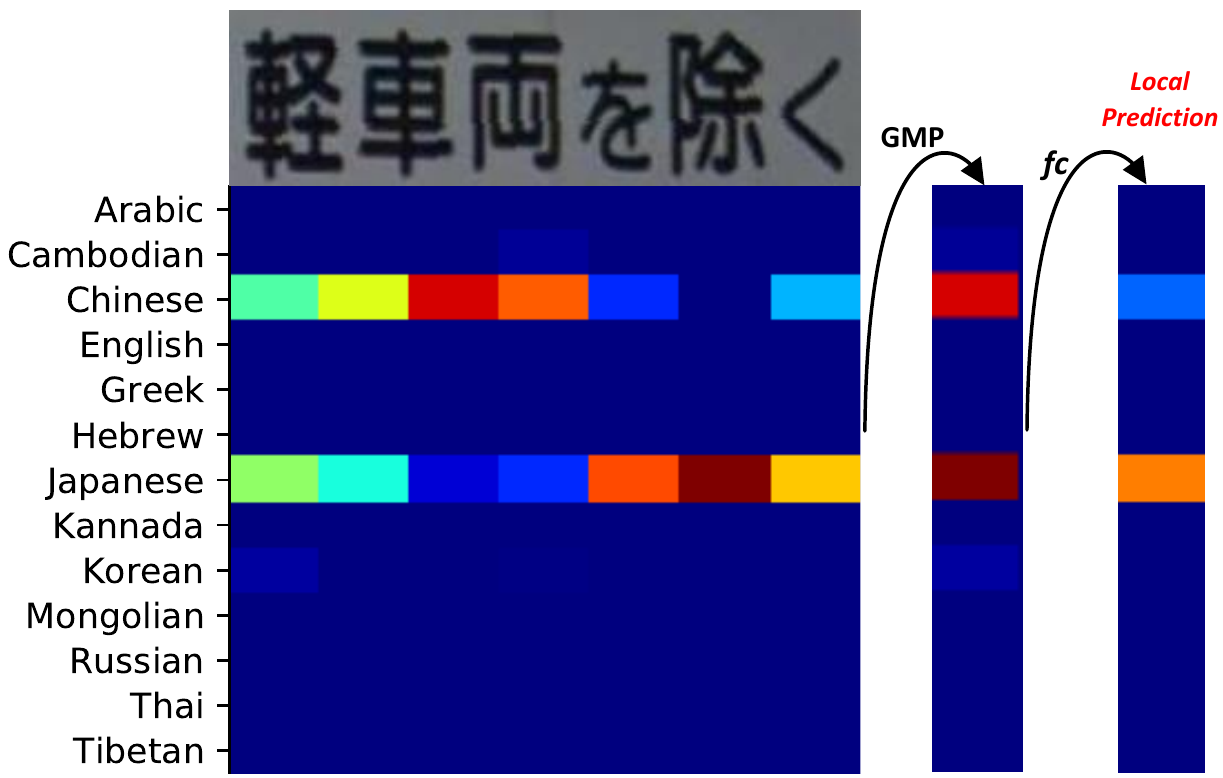}
\label{visa}}
\subfigure[]{\includegraphics[scale=0.61]{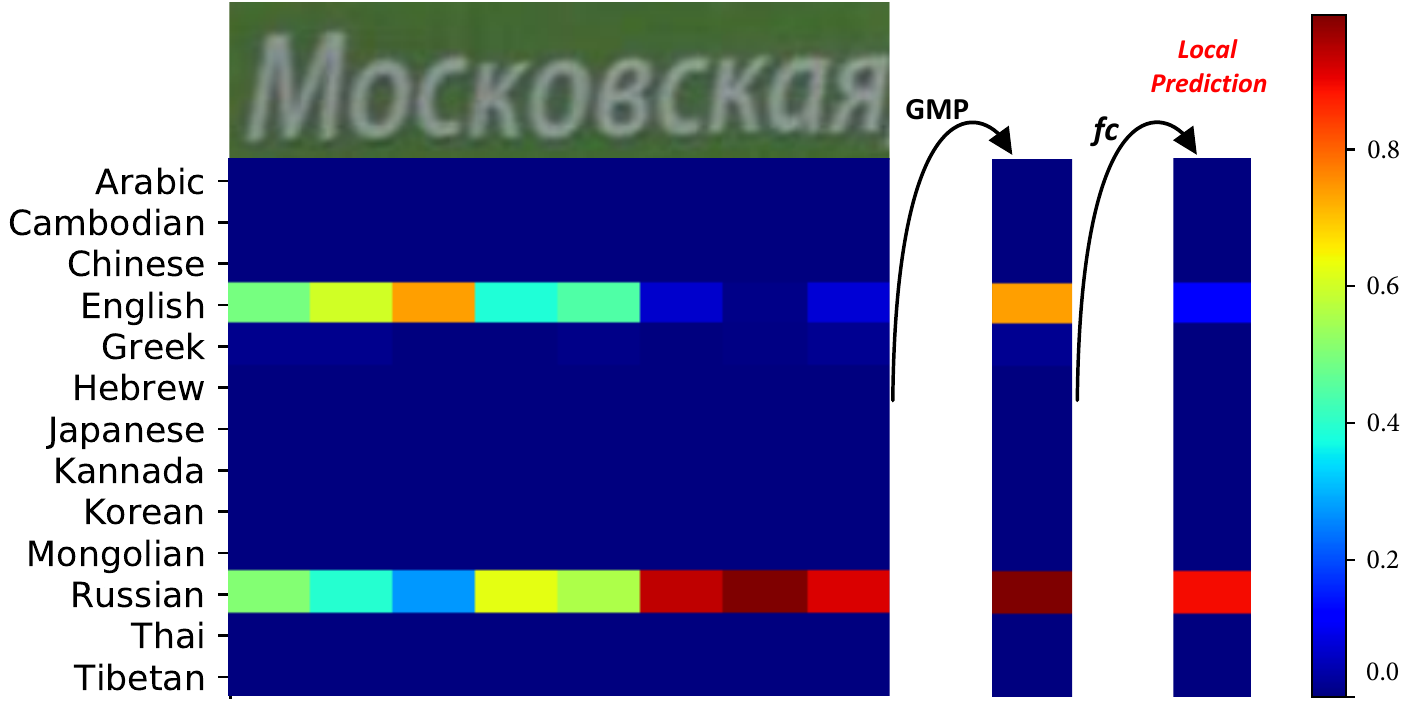}
\label{visb}}
}
\caption{Visualization of the behaviour of PA in our experiments. Two difficult test samples are taken as examples. Note that there actually exists overlap between patch neighbors. (a)Some patches are likely to be Chinese, while some are Japanese. We can regard the two scripts as abstract high-level semantic features by performing GMP and make inference for further local prediction. So we can assert it is Japanese instead of Chinese. (b) The same logical flow as in (a). English and Russian are potentially the two high-level semantic features, and PA can accurately identify it as Russian.
\label{vis}}
\vspace*{-0.5cm}
\end{figure*}

As for scene text line images in SIW-13, a great improvement has been made with a good balance among all scripts. The Sequence-to-label problem demands more comprehensive feature representation than sequential dependency, which is proved by the comparison between ours and CRNN~\cite{CRNN} which has a popular application in scene text recognition~\cite{shi2017end}. Besides, Mei~\cite{CRNN} cost much time to predict a text line which may be caused by the sequential computation in RNN.
Zdenek~\cite{bag} used BLCT to enhance the discriminative features. But they imposed $idf$(inverse document frequency) to get codewords occurrences, suffering a lot from the impairment caused by less critical features. The image in \fig{visa} which has much more Chinese patches than kana was misclassified to Chinese by their model. Bhunia~\cite{attLSTM} coupled local and global features, but it suffers from the impairment too. What is more, the use of many cropped patches can make considerably redundant computation and memory usage which can influence the efficiency especially in its LSTM module which precludes parallelization.
Our model built with 26.7 MB parameters takes about 2.5 ms per image when we make test one by one, owing to the efficient matrix computation with a full-size image and the simple pipeline.

For CVSI2015, where the images with single background occurring in video caption, our method can reach the best accuracy among the published works. Former works are usually not able to have a proper balance between scene text and video caption, like Shi~\cite{wildSI} and Nicolaou~\cite{nicolaou2016visual}.

Our method also achieves the best performance on RRC-MLT2017, pushing itself to more complex scenarios and to be more practical. Besides the results shown in \tab{results}, Bhunia~\cite{attLSTM} conducted their model on the validation set and achieved 90.23\%, while our approach can reach up to 95.31\% on it.

\subsection{Ablation Study}
We conduct ablation studies on SIW-13 to show the power of our proposed PA along with GS and softermax loss.

\subsubsection{The contribution of the proposed module}
Here we consider the contribution of PA by replacing the two-module (GS and PA) parts with other modules alternately while keeping the shared feature extractor.

\tab{branch} shows the results in detail, where \textbf{GS} means a single GS module is used without PA, and \textbf{PA} has the similar meaning. \textbf{GS+GS} is an ensemble model which can be regarded as that another GS takes the place of PA in \fig{fig3}. \textbf{GS+GMP} just changes the GAP operation into GMP in one of the modules in \textbf{GS+GS}. \textbf{GS+PA} is the exact proposed method.

A single module is not enough for a fine-grained classifier to exploit information both globally and locally, which can be reflected by the result of \textbf{GS} and \textbf{PA}. GS cannot notice the fine-grained detail well and PA is prone to be limited in a sub-area. The ensemble model \textbf{GS+GS} only gets a slight improvement compared with \textbf{GS}, turning out that our proposed method should not attribute its great performance to ensemble. \textbf{GS+GMP} use GMP to extract the most remarkable responses in 512 dimensions which can be regarded as a kind of local features obtained in another way, but every dimension do not has an explicit meaning and cannot be supervised by label. So it only improves 0.3\%, yet holding much more parameters. All of them highlight our proposed method, showing the power of integration of GS and PA.

\subsubsection{Effect of softermax loss}
The proposed softermax loss mentioned in \ref{softer} is vital for PA in the training stage. We have investigated whether the supervision works and the importance of softermax loss.

Ablation results are shown in \tab{supervision}. The intermediate supervision makes great sense on the final accuracy, which can guide the mid-prediction to reach our expectation with explicit meaning. The weight $\lambda$ in \eq{L4} determines the influence of the softermax loss. ``$\lambda=0$" in \tab{supervision} means that only softmax loss conducts the supervision. The result shows the significance of the softerness brought by softermax loss.

\subsection{Visualization Analysis}\label{visualization}
Insights into the behaviour of our proposed PA can be obtained by visualizing the vectors with $\mathrm{K}$ dimensions which are probability distributions over $\mathrm{K}$ classes. Specifically, we take the patch-level prediction, vector after GMP and the local prediction from the linear classifier (fc) as the objects to observe.

As shown in \fig{vis}, predictions for patches are not forced to an extreme and the probabilities scatter relatively high over several(here is 3) scripts, which agrees with the fact that a patch alone regarded as an independent subsample from input can actually correspond to several scripts.
By GMP, a vector consisting of the most remarkable response over classes is actually a kind of high-level semantic feature which shows what scripts the components of the input could be. The local-level prediction can be obtained by further inference which is exactly a simple linear classifier. We can make full use of fine-grained discriminative features through the proposed procedure.

\section{Conclusion}
We present a simple but effective approach for scene text script identification. Patch Aggregator can learn discriminative features while having discriminatory power not been reduced by redundant features. It significantly improves the baseline model, Global Squeezer. The novel softermax loss is designed to make intermediate supervision on patch-level prediction. Our method achieves the best results on three benchmark datasets, demonstrating its great power.

\section*{Acknowledgment}
This research was supported by the National Natural Science Foundation of China (NSFC) grants 61733007 and 61773176. Dr. Xiang Bai was supported by the National Program for Support of Top-notch Young Professionals and the Program for HUST Academic Frontier Youth Team.

\bibliographystyle{IEEEtran}
\tiny
\bibliography{SI}

\end{document}